\documentclass[conference]{IEEEtran}
\IEEEoverridecommandlockouts
\usepackage{cite}
\usepackage{amsmath,amssymb,amsfonts}
\usepackage{algorithmic}
\usepackage{graphicx}
\usepackage{textcomp}
\usepackage{xcolor}
\usepackage[hidelinks]{hyperref}
\usepackage{makecell}
\usepackage{multirow}
\usepackage{subcaption}
\usepackage{booktabs}
\usepackage{listings}
\usepackage{amsmath}
\usepackage{amssymb}

\definecolor{ps-brown}{rgb}{0.65, 0.16, 0.16}
\definecolor{ps-orange}{rgb}{1.0, 0.65, 0}
\definecolor{ps-white}{rgb}{1, 1, 1}
\definecolor{ps-blue}{rgb}{0, 0, 1}
\definecolor{ps-commentblue}{rgb}{0.13725490196078433, 0.4549019607843137, 0.8666666666666667}
\definecolor{ps-yellow}{rgb}{1, 1, 0}
\definecolor{ps-darkred}{rgb}{0.55, 0, 0}
\definecolor{ps-red}{rgb}{1, 0, 0}
\definecolor{ps-lightred}{rgb}{1, 0.41, 0.41}
\definecolor{ps-green}{rgb}{0, 1, 0}
\definecolor{ps-black}{rgb}{0, 0, 0}
\definecolor{ps-keywordpurple}{rgb}{0.564, 0.427, 0.843}
\definecolor{ps-purple}{rgb}{0.6901960784313725, 0.11372549019607843, 0.7568627450980392} 
\definecolor{ps-background}{rgb}{0.058823529411764705, 0.09803921568627451, 0.16470588235294117}

\lstdefinestyle{puzzlescript}{
    backgroundcolor=\color{ps-background},   
    basicstyle=\ttfamily\footnotesize\color{white},
    keywordstyle={\color{ps-keywordpurple}},
    commentstyle=\color{ps-commentblue},
    stringstyle=\color{ps-orange},
    numbers=none,
    frame=none,
    framerule=0pt,
    framexleftmargin=0em,
    framexrightmargin=0em,
    framextopmargin=0pt,
    framexbottommargin=0pt,
    xleftmargin=0em,
    xrightmargin=0em,
    breaklines=false,
    breakatwhitespace=false,
    showspaces=false,
    showstringspaces=false,
    showtabs=false,
    columns=flexible,
    keepspaces=true,
    moredelim=[is][\color{gray}]{startlevel}{endlevel},
    morecomment=[s]{(}{)},
    literate={->}{{{\color{ps-red}{->}}}}2
             {>}{{{\color{ps-purple}{>}}}}1
             {<}{{{\color{ps-purple}{<}}}}1
             {\\wedge}{{\color{ps-purple}{$\wedge${}}}}3
             {\\vee}{{{\color{ps-purple}{$\vee$}}}}4,
    keywords={title, author, OBJECTS, LEGEND, SOUNDS, COLLISIONLAYERS, RULES, WINCONDITIONS, all, on, LEVELS},
    keywords=[2]{Background, Player, Box, Trigger, Switch, GateClosed, GateOpen, Wall, Toggle},
    keywordstyle=[2]{\color{ps-green}},
    keywords=[3]{Unconventional, PushPull, ChatGPT},
    keywordstyle=[3]{\color{ps-orange}},
    keywords=[4]{late, no},
    keywordstyle=[4]{\color{ps-purple}}
}

\def\BibTeX{{\rm B\kern-.05em{\sc i\kern-.025em b}\kern-.08em
    T\kern-.1667em\lower.7ex\hbox{E}\kern-.125emX}}
\begin{document}

\title{ScriptDoctor: Automatic Generation of PuzzleScript Games via Large Language Models and Tree Search}



\author{
\IEEEauthorblockN{Sam Earle\IEEEauthorrefmark{1}, 
Ahmed Khalifa\IEEEauthorrefmark{2}, 
Muhammad Umair Nasir\IEEEauthorrefmark{3}, 
Zehua Jiang\IEEEauthorrefmark{1}, 
Graham Todd\IEEEauthorrefmark{1}, 
Andrzej Banburski-Fahey\IEEEauthorrefmark{4}, \\
Julian Togelius\IEEEauthorrefmark{1}}
\IEEEauthorblockA{\IEEEauthorrefmark{1}New York University, Brooklyn, USA\\
\{sam.earle, zj2086, gdt9380\}@nyu.edu, julian@togelius.com%
}
\IEEEauthorblockA{\IEEEauthorrefmark{2}University of Malta, Msida, Malta\\ahmed@khalifa.com}%
\IEEEauthorblockA{\IEEEauthorrefmark{3}University of the Witwatersrand, Johannesburg, South Africa.\\muhammad.nasir@witz.ac.za}
\IEEEauthorblockA{\IEEEauthorrefmark{4}Microsoft, Redmond, USA.\\abanburski@microsoft.com}%
}

\maketitle

\begin{abstract}
There is much interest in using large pre-trained models in Automatic Game Design (AGD), whether via the generation of code, assets, or more abstract conceptualization of design ideas. But so far this interest largely stems from the ad hoc use of such generative models under persistent human supervision. Much work remains to show how these tools can be integrated into longer-time-horizon AGD pipelines, in which systems interface with game engines to test generated content autonomously. To this end, we introduce ScriptDoctor, a Large Language Model (LLM)-driven system for automatically generating and testing games in PuzzleScript, an expressive but highly constrained description language for turn-based puzzle games over 2D gridworlds. ScriptDoctor generates and tests game design ideas in an iterative loop, where human-authored examples are used to ground the system's output, compilation errors from the PuzzleScript engine are used to elicit functional code, and search-based agents play-test generated games. ScriptDoctor serves as a concrete example of the potential of automated, open-ended LLM-based workflows in generating novel game content.
\end{abstract}


\section{Introduction}

Can large language models (LLMs) create complete games? Spend some time scrolling through the right social media feeds, and you will see ample examples of this happening. So, apparently, yes, they can. Yet, many questions are hard to answer. How good are these games? We do not yet have the capabilities to play and test games in general, so this requires human evaluation. With the very large number of small games in the training set of a modern LLM, there remains the suspicion that the generated games are at best a minor variation on something it has already seen. The lack of automated evaluation is also limiting our ability to understand \emph{how} to make an LLM produce the games we want.

In this paper, we seek to shed light on some of these questions not by investigating unconstrained game generation in JavaScript or Python, but by using a model organism of sorts. We investigate LLM-based game generation in a domain-specific language that allows for the creation of complete games in a manageable amount of text, which allows for generating diverse high-quality games, which allows for automatic game-testing through AI-based game-playing, and for which we have an approximate understanding of which games have ever been made public. Our model organism is \emph{PuzzleScript}~\footnote{\url{https://www.puzzlescript.net/}}, a domain-specific language for puzzle games. This has multiple advantages. PuzzleScript was created as a way of prototyping puzzle games, and has since been used by hundreds of game designers to create thousands of games, some of them quite good and not all of them puzzle games. Importantly, the PuzzleScript engine has a standard format for input and output, and is lightweight enough to allow rapid rollouts.


\begin{figure*}
    \centering
    \includegraphics[width=1.0\linewidth]{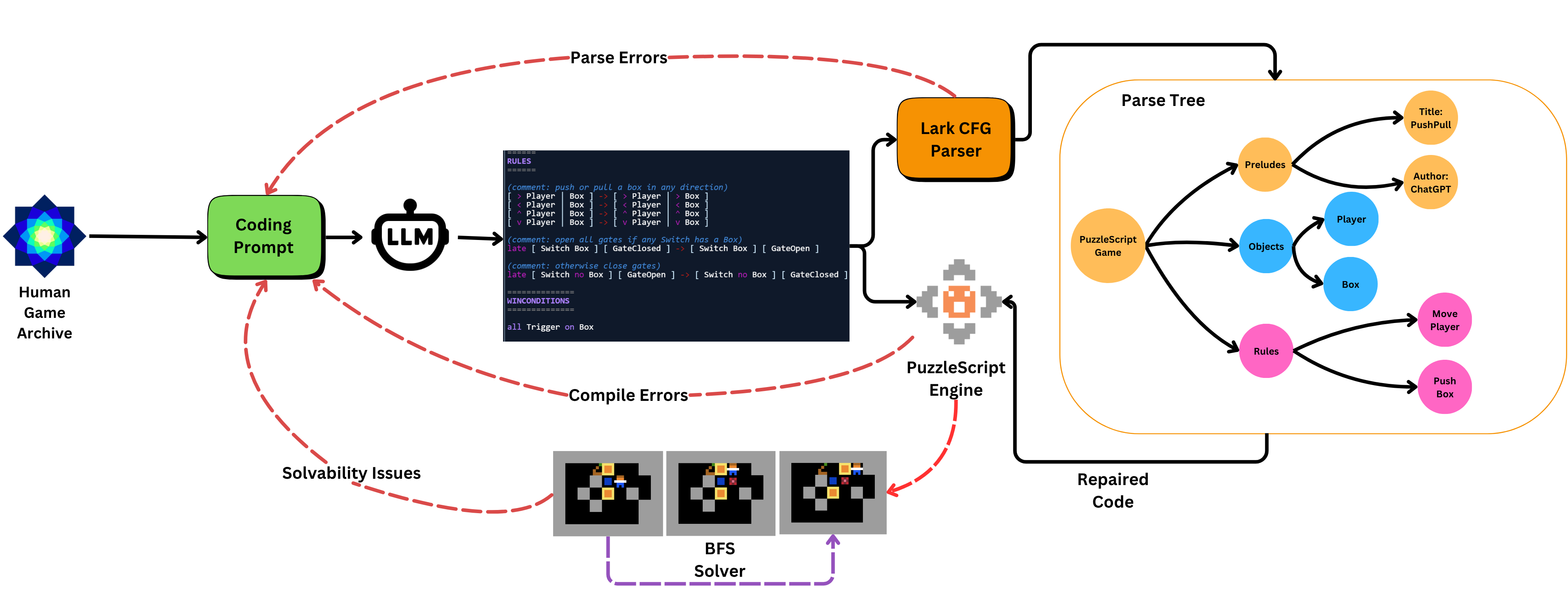}
    \caption{\textbf{The ScriptDoctor automatic game generation pipeline.} An LLM iterates on PuzzleScript code given feedback from the compiler and a search-based player agent. Its output is parsed and repaired where possible according to a context-free grammar, and its prompt is augmented by documentation, examples of human games, and design ideas generated by a ``brainstorming'' agent.}
    \label{fig:pipeline}
\end{figure*}

\section{Related work}

Generating game rules or complete games is relatively uncommon in PCG work~\cite{yannakakis2018artificial}. This is due to the complexity of the task and the difficulty of algorithmically evaluating overall game quality. Most game generation is based on the search-based paradigm~\cite{togelius2011search} using playability information yielded from simulation and/or expert knowledge. ~\cite{browne2008automatic} creates novel board games while ~\cite{togelius2008experiment} generates rules for Pac-Man-like games. A number of subsequent papers describe attempts to generate game rules or full games using similar ideas~\cite{treanor2012game,cook2013mechanic,nielsen2015towards,khalifa2017general,gonzalez2023mechanic}.

Particularly relevant to this paper is \cite{khalifa2015automatic}, which generates PuzzleScript games. They fixed the number of game objects, and used a genetic algorithm to evolve a small subset of game rules. They use a constructive level generator that takes the generated rules and uses expert knowledge to place objects. The final levels and rules are evaluated using an automated player and compared to expert knowledge about the domain and game player data. The authors argue that having a better orchestration~\cite{liapis2018orchestrating} between levels and rules using co-evolution~\cite{ma2018survey} might improve the results. Even so, the possibility space of generated games is limited by dependency on hand-crafted constructive algorithms.

With the rise of LLMs~\cite{zhao2023survey}, one might think that the models would have some heuristics, perhaps implicit, for what makes a game good. This pushed researchers to explore the capabilities of these models in games~\cite{gallotta2024large}. \cite{hu2024game} few-shot prompt LLMs to generate a simple maze game for the General Video Game framework~\cite{perez2019general}. \cite{abagames2024claude} leverage LLMs' understanding of programming languages to generate arcade games written in a simple JavaScript game engine. \cite{todd2024gavel} mix search-based methods~\cite{togelius2011search} with LLMs to generate new games in the Ludii framework using a fine-tuned model.

This work sets itself apart from previous ones as it is not just producing game rules but also levels, and to some extent, visual aesthetics and narrative. We attempt to give LLMs the power to orchestrate between several different facets of the game~\cite{liapis2018orchestrating} and produce a full PuzzleScript game in an unconstrained fashion. 
While this work investigates how the pre-trained knowledge of LLMs can be leveraged out of the box to support designers/developers, future work might investigate fine-tuning LLMs for this purpose, as in GAVEL~\cite{todd2024gavel}.

\section{Methods}
In this section, we outline ScriptDoctor, a pipeline for the automatic, iterative generation of entire PuzzleScript games that leverages feedback from a tree search-based player agent.
A schematic overview of the system is depicted in \ref{fig:pipeline}.

Our system is developed as a fork of the original PuzzleScript repository~\footnote{\url{https://github.com/increpare/PuzzleScript}}. We communicate with the existing game engine via a Python server. The server is responsible for controlling the hyperparameters of the game generation process and querying LLMs.

On the client side, we compile the generated PuzzleScript games, and compilation errors or warnings from the engine are returned as potential feedback for the LLM. If the game compiles, each level is tested for playability using a breadth-first search algorithm. The algorithm runs until a solution is found or it explores 1M unique nodes. The results (solvability, number of expanded nodes, and solution length) from the BFS algorithm are sent back to the server to help with the generation process.

The language model is given $10$ total chances to output functional code.
A generated script is deemed successful when it compiles successfully and each of its levels admits a solution of length $>10$.
At this point, the game generation trial is terminated.
Otherwise, the LLM is prompted with the code that it generated at the prior iteration, any compilation warnings and errors that appeared in the PuzzleScript console, and feedback from the solver. We also define PuzzleScript as a Context-Free Grammar (CFG) using the Lark library for Python~\footnote{\url{https://github.com/lark-parser/lark}}, and additionally include any syntax errors identified by this processing as feedback to the puzzle-generating LLM.

\section{Experiments}
We experimented with generating games with zero-shot and few-shot prompting. In few-shot prompting, we sampled a group of random human-made games from a small dataset ($610$) of PuzzleScript games. The sampling continues until we reach the LLM context size, and adding another game will surpass the maximum number of tokens. This dataset is scraped from the archive at ~\cite{pedro2019database}. We also experimented with chain-of-thought reasoning to see the effect of this technique on the generated games. We tested different LLM models (GPT-4o~\cite{gpt4o}, o1~\cite{jaech2024openai}, and o3-mini~\cite{o3-mini}) with varying context length ($10,000$, $30,000$, $50,000$, and $70,000$). 

To asses the generated games, we measure the percentage of games that compile successfully in the PuzzleScript editor. We also measure the percentage of generated games wherein \textbf{any} level admits a solution of $>1$ moves, and the percentage of generated games wherein \textbf{all} levels admit solutions of $>10$ moves. (We note for context that many human-authored games admit short solutions while still being interesting or challenging.) To assess the \textbf{complexity} of generated solutions, we consider the number of nodes explored by the BFS player agent.

\section{Results}

\begin{table}
\begin{center}
\caption{Effect of few-shot prompting GPT-4o by including a random subset of high-quality human-authored games in the prompt during code generation and repair (up to 30k tokens worth). Results aggregated over 20 game generation trials.}
\begin{tabular}{llllllllll}
\toprule
 &  & Compiles & \makecell{Any\\Solvable} & \makecell{All\\Solvable} & Sol. Complexity \\
Fewshot & CoT &  &  &  &  \\
\midrule
\multirow[t]{2}{*}{F} & F & 30\% & 0\% & 0\% & 0 ± 0 \\
 & T & 25\% & 10\% & \textbf{5\%} & 13 ± 45 \\
\cline{1-6}
\multirow[t]{2}{*}{T} & F & 70\% & 40\% & 0\% & \textbf{1,709} ± 6,722  \\
 & T & \textbf{80\%} & \textbf{55\%} & \textbf{5\%} & 1,498 ± 6,294 \\
\bottomrule
\end{tabular}

\label{tab:fewshot_21}
\end{center}
\end{table}

\begin{table}
\begin{center}
\caption{Comparison between models, each generating $10$ games using fewshot prompting and a context length of 10k tokens.}
\begin{tabular}{lllllllll}
\toprule
 & Compiles & \makecell{Any\\Solvable} & \makecell{All\\Solvable} & Sol. Complexity \\
Model &  &  &  &   \\
\midrule
GPT-4o & 67\% & 40\% & \textbf{7\%} & 19 ± 27 \\
o1 & 87\% & \textbf{67\%} & \textbf{7\%} & \textbf{22,771} ± 84,485 \\
o3-mini & \textbf{93\%} & 47\% & 20\% & 2,676 ± 9,888  \\
\bottomrule
\end{tabular}

\label{tab:models}
\end{center}
\end{table}


In \autoref{tab:fewshot_21}, we observe the effect of few-shot prompting GPT-4o by including human-authored games in the model's system prompt and chain-of-thought prompting it. We see that few-shot prompting increases the quality of generated code in terms of the proportion of generated games that successfully compile and the percentage of games that are solvable via tree search. It likewise results in increased complexity of solutions to generated games.

\begin{figure*}
\begin{subfigure}{0.195\linewidth}
\centering
\includegraphics[width=\linewidth]{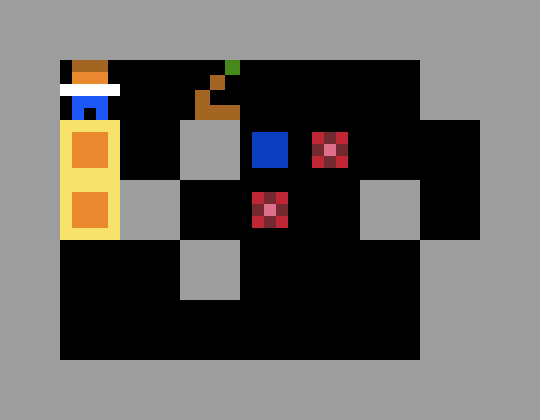}
\end{subfigure}
\hfill
\begin{subfigure}{0.195\linewidth}
\centering
\includegraphics[width=\linewidth]{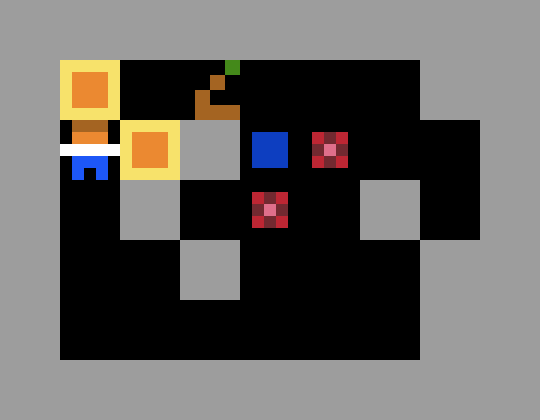}
\end{subfigure}
\hfill
\begin{subfigure}{0.195\linewidth}
\centering
\includegraphics[width=\linewidth]{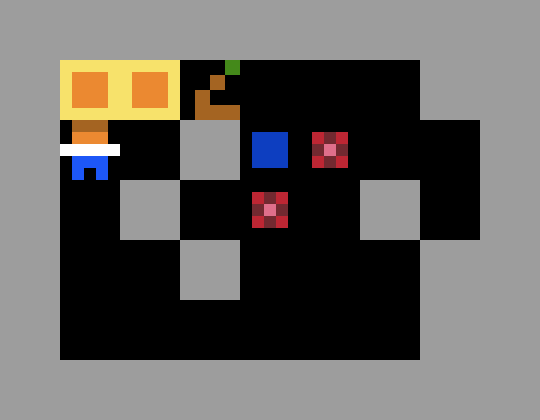}
\end{subfigure}
\hfill
\begin{subfigure}{0.195\linewidth}
\centering
\includegraphics[width=\linewidth]{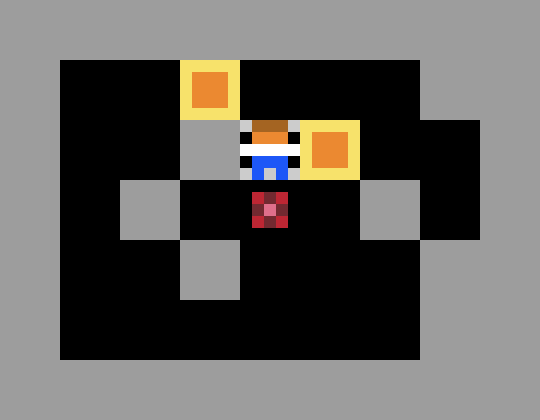}
\end{subfigure}
\hfill
\begin{subfigure}{0.195\linewidth}
\centering
\includegraphics[width=\linewidth]{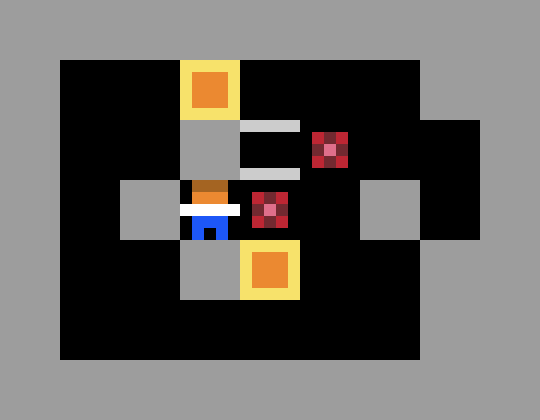}
\end{subfigure}
\caption{Select frames from a solution to the \textit{Unconventional PushPull} puzzle generated by ScriptDoctor with o1. The player must place crates (gold boxes) on targets (red squares), by pushing and pulling them, as well as ``sliding'' them one tile beyond the player's immediate neighborhood when the player is obstructed in the relevant direction. The switch (brown and green lever) can be activated by crates in order to make the gate (blue) traversible. The most direct solution is 34 moves long and takes breadth-first search 1,006 iterations to discover. To a human player, the solution is tricky but sensible. For the sake of visualization, we replace generated sprites (which in this case are abstract and sometimes invisible) with semantically similar sprites from our dataset of human-authored games.}
\label{fig:pushpull}
\end{figure*}

In \autoref{tab:models}, we compare different LLM models that are used in ScriptDoctor and find that reasoning models o1 and o3-mini outperform GPT-4o in terms of the functionality and complexity of generated games. (This aligns with the improvement elicited by chain-of-thought prompting in \autoref{tab:fewshot_21}.)
A particularly challenging mechanic and level design by o1 is featured in \autoref{fig:pushpull}

We also compare the effect of different context lengths on GPT-4o in \autoref{tab:ctx}. We notice that increasing context length allows the system to generate more compilable games, which might be due to the effect of having more games in the prompt. We also notice that the quality doesn't change that drastically after $30,000$ tokens. We believe that after a certain point of context length, having more games doesn't add much to the model's understanding of what it needs to create. Future work might be needed to investigate what can be provided to improve the generated games besides adding few-shot examples.

\begin{table}
\begin{center}
\caption{Effect of increasing GPT-4o's context window when including human-authored games in the prompt.}
\begin{tabular}{lllllllll}
\toprule
 & Compiles & \makecell{Any\\Solvable} & \makecell{All\\Solvable} & Sol. Complexity \\
Context Length &  &  &  &   \\
\midrule
10,000 & 60\% & 30\% & \textbf{10\%} & 18 ± 28\\
30,000 & \textbf{100\%} & \textbf{80\%} & 0\% & 46 ± 44 \\
50,000 & 90\% & 40\% & \textbf{10\%} & \textbf{7,003} ± 19,329 \\
70,000 & \textbf{100\%} & 40\% & 0\% & 203 ± 553 \\
\bottomrule
\end{tabular}

\label{tab:ctx}
\end{center}
\end{table}



\section{Discussion}

The strong positive effect of a few-shot prompting the system with human games is unsurprising. PuzzleScript is not nearly as prevalent as a coding language, such as Python or C++, so having human examples is crucial to facilitate the model's outputting syntactically correct and compilable code.

Without finetuning, LLMs struggle with the kind of spatial reasoning at work in level generation~\cite{sudhakaran2024mariogpt,todd2023level}.
Often, ScriptDoctor outputs levels that admit overly short solutions.
When asked to complexify their solutions, it attempts to edit the level layout to make it more challenging, but often this only amounts to stretching the level out by a few rows or columns, or adding some scattered obstacles.

The output of our system may serve as a dataset for fine-tuning smaller, more accessible LLMs. Given that our pipeline verifies the compilability and solvability of generated games, it could provide a suitable dataset for supervised fine-tuning such models (alongside the dataset of human-authored games). Having a fine-tuned model would allow us to perform constrained generation of PuzzleScript code using our Context-Free Grammar specification of the PuzzleScript language, effectively guaranteeing the syntactical correctness of generated code similar to work by Todd et al.~\cite{todd2024gavel}.

\begin{figure}
\begin{subfigure}{0.325\linewidth}
\centering
\includegraphics[width=\linewidth]{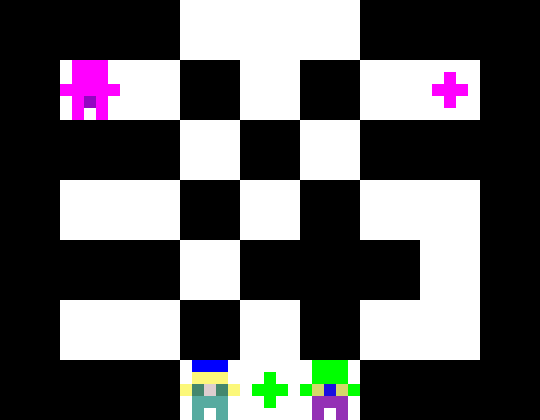}
\end{subfigure}
\hfill
\begin{subfigure}{0.325\linewidth}
\centering
\includegraphics[width=\linewidth]{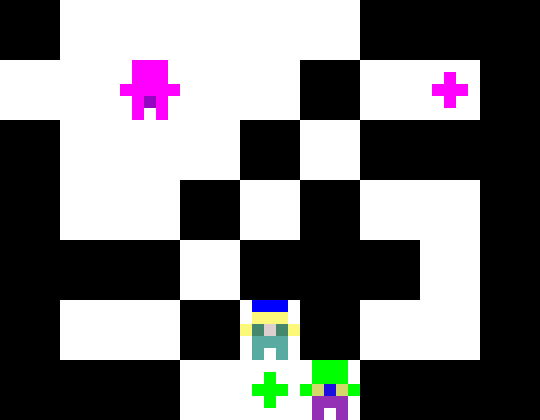}
\end{subfigure}
\hfill
\begin{subfigure}{0.325\linewidth}
\centering
\includegraphics[width=\linewidth]{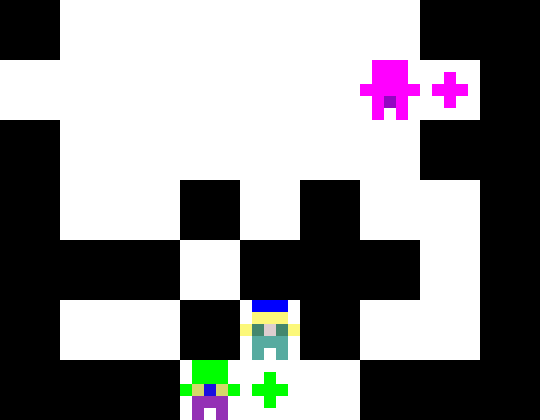}
\end{subfigure}
\caption{Select frames from a level of a generated game by GPT-4o. This level involves simultaneous control of $2$ characters. The wizard (pink) can teleport through walls if they aren't adjacent or destroy them otherwise without moving. Taking advantage of the remove mechanic, the solver can reposition the rogue (green) such that one final move to the right will result in both the wizard and the rogue reaching their targets.}
\label{fig:trio_quest}
\end{figure}

While information about solvability is valuable for an iterative code-generating LLM, it may well be insufficient for capturing whether mechanics are ``broken'' relative to the model's intent.
In fact, we observe that the most complex games tend to be solvable or complex in spite of or even as a result of their broken mechanics (e.g. \autoref{fig:trio_quest}). For the system to reliably fix such issues, some more targeted form of feedback is likely necessary.
One option could be to collect frames of gameplay in which particular rules were applied, and supply a vision-language model with the appropriate context to be able to diagnose potential flaws in their implementation.

\section{Conclusion}

Despite increasing interest in generating arbitrary video games with LLMs via quasi-passive ``vibe coding'', there exist few means for quantitatively evaluating these models' ability to produce novel and functional games autonomously. We answer the call with ScriptDoctor, an automatic game designer that integrates LLMs with search-based player agents and grammatical parse trees to generate concise but expressive games in PuzzleScript. The system operates in a possibility space that is both open-ended---with generated scripts defining everything from sprites, to mechanics, to levels---and tractable for artificial playtesting---with a small, fixed action space.

We find that using few-shot prompting with examples of human-authored games increases the functionality of generated code. We find that frontier reasoning models o1 and o3-mini are more capable than their non-reasoning counterpart GPT-4o in generating compilable and playable games. Increasing the context length to include more human-authored games shows diminishing returns after $30,000$ tokens, suggesting that the model might need more information to understand how to reason beyond these examples.

Future work can leverage ScriptDoctor's automatic metrics of quality to guide the generative process, wrapping it in a novelty-seeking evolutionary loop. And PuzzleScript's pattern rewrite rules lend themselves both to static causal analysis and reimplementation via GPU-compatible convolutions for accelerated playtesting---both potentially valuable signals during the open-ended search for novel video games.

\bibliography{ref}

\begin{thebibliography}{10}
\providecommand{\url}[1]{#1}
\csname url@samestyle\endcsname
\providecommand{\newblock}{\relax}
\providecommand{\bibinfo}[2]{#2}
\providecommand{\BIBentrySTDinterwordspacing}{\spaceskip=0pt\relax}
\providecommand{\BIBentryALTinterwordstretchfactor}{4}
\providecommand{\BIBentryALTinterwordspacing}{\spaceskip=\fontdimen2\font plus
\BIBentryALTinterwordstretchfactor\fontdimen3\font minus
  \fontdimen4\font\relax}
\providecommand{\BIBforeignlanguage}[2]{{%
\expandafter\ifx\csname l@#1\endcsname\relax
\typeout{** WARNING: IEEEtran.bst: No hyphenation pattern has been}%
\typeout{** loaded for the language `#1'. Using the pattern for}%
\typeout{** the default language instead.}%
\else
\language=\csname l@#1\endcsname
\fi
#2}}
\providecommand{\BIBdecl}{\relax}
\BIBdecl

\bibitem{yannakakis2018artificial}
G.~N. Yannakakis and J.~Togelius, \emph{Artificial intelligence and
  games}.\hskip 1em plus 0.5em minus 0.4em\relax Springer, 2018, vol.~2.

\bibitem{togelius2011search}
J.~Togelius, G.~N. Yannakakis, K.~O. Stanley, and C.~Browne, ``Search-based
  procedural content generation: A taxonomy and survey,'' \emph{Transactions on
  Computational Intelligence and AI in Games}, vol.~3, no.~3, 2011.

\bibitem{browne2008automatic}
C.~B. Browne, ``Automatic generation and evaluation of recombination games,''
  Ph.D. dissertation, Queensland University of Technology, 2008.

\bibitem{togelius2008experiment}
J.~Togelius and J.~Schmidhuber, ``An experiment in automatic game design,'' in
  \emph{Symposium On Computational Intelligence and Games}.\hskip 1em plus
  0.5em minus 0.4em\relax IEEE, 2008.

\bibitem{treanor2012game}
M.~Treanor, B.~Blackford, M.~Mateas, and I.~Bogost, ``Game-o-matic: Generating
  videogames that represent ideas,'' in \emph{Procedural Content Generation in
  Games Workshop}, 2012.

\bibitem{cook2013mechanic}
M.~Cook, S.~Colton, A.~Raad, and J.~Gow, ``Mechanic miner: Reflection-driven
  game mechanic discovery and level design,'' in \emph{EvoApplications, EvoStar
  Conference}.\hskip 1em plus 0.5em minus 0.4em\relax Springer, 2013.

\bibitem{nielsen2015towards}
T.~S. Nielsen, G.~A. Barros, J.~Togelius, and M.~J. Nelson, ``Towards
  generating arcade game rules with vgdl,'' in \emph{Computational Intelligence
  and Games Conference}.\hskip 1em plus 0.5em minus 0.4em\relax IEEE, 2015.

\bibitem{khalifa2017general}
A.~Khalifa, M.~C. Green, D.~Perez-Liebana, and J.~Togelius, ``General video
  game rule generation,'' in \emph{Computational Intelligence and Games
  Conference}.\hskip 1em plus 0.5em minus 0.4em\relax IEEE, 2017.

\bibitem{gonzalez2023mechanic}
J.~J. Gonzalez, S.~Cooper, and M.~Guzdial, ``Mechanic maker 2.0: reinforcement
  learning for evaluating generated rules,'' in \emph{Artificial Intelligence
  and Interactive Digital Entertainment Conference}, vol.~19, no.~1, 2023.

\bibitem{khalifa2015automatic}
A.~Khalifa and M.~Fayek, ``Automatic puzzle level generation: A general
  approach using a description language.''\hskip 1em plus 0.5em minus
  0.4em\relax IEEE, 2015.

\bibitem{liapis2018orchestrating}
A.~Liapis, G.~N. Yannakakis, M.~J. Nelson, M.~Preuss, and R.~Bidarra,
  ``Orchestrating game generation,'' \emph{Transactions on Games}, vol.~11,
  no.~1, 2018.

\bibitem{ma2018survey}
X.~Ma, X.~Li, Q.~Zhang, K.~Tang, Z.~Liang, W.~Xie, and Z.~Zhu, ``A survey on
  cooperative co-evolutionary algorithms,'' \emph{Transactions on Evolutionary
  Computation}, vol.~23, no.~3, 2018.

\bibitem{zhao2023survey}
W.~X. Zhao, K.~Zhou, J.~Li, T.~Tang, X.~Wang, Y.~Hou, Y.~Min, B.~Zhang,
  J.~Zhang, Z.~Dong \emph{et~al.}, ``A survey of large language models,''
  \emph{arXiv preprint arXiv:2303.18223}, 2023.

\bibitem{gallotta2024large}
R.~Gallotta, G.~Todd, M.~Zammit, S.~Earle, A.~Liapis, J.~Togelius, and G.~N.
  Yannakakis, ``Large language models and games: A survey and roadmap,''
  \emph{arXiv preprint arXiv:2402.18659}, 2024.

\bibitem{hu2024game}
C.~Hu, Y.~Zhao, and J.~Liu, ``Game generation via large language models,'' in
  \emph{Conference on Games}.\hskip 1em plus 0.5em minus 0.4em\relax IEEE,
  2024.

\bibitem{perez2019general}
D.~Perez-Liebana, J.~Liu, A.~Khalifa, R.~D. Gaina, J.~Togelius, and S.~M.
  Lucas, ``General video game ai: A multitrack framework for evaluating agents,
  games, and content generation algorithms,'' \emph{Transactions on Games},
  vol.~11, no.~3, 2019.

\bibitem{abagames2024claude}
K.~Cho, ``Can ai chatbots create new games?''
  \url{https://abagames.github.io/joys-of-small-game-development-en/generation/can_ai_chatbot_create_game.html},
  2024, accessed on: October 23, 2024.

\bibitem{todd2024gavel}
G.~Todd, A.~Padula, M.~Stephenson, {\'E}.~Piette, D.~J. Soemers, and
  J.~Togelius, ``Gavel: Generating games via evolution and language models,''
  \emph{arXiv preprint arXiv:2407.09388}, 2024.

\bibitem{pedro2019database}
\BIBentryALTinterwordspacing
Pedro, ``Puzzlescript games database,'' 2019, accessed: March 14, 2025.
  [Online]. Available: \url{https://pedros.works/puzzlescript/database}
\BIBentrySTDinterwordspacing

\bibitem{gpt4o}
\BIBentryALTinterwordspacing
OpenAI, ``Hello gpt-4o,'' May 2024, accessed: 2024-09-11. [Online]. Available:
  \url{https://openai.com/index/hello-gpt-4o/}
\BIBentrySTDinterwordspacing

\bibitem{jaech2024openai}
A.~Jaech, A.~Kalai, A.~Lerer, A.~Richardson, A.~El-Kishky, A.~Low, A.~Helyar,
  A.~Madry, A.~Beutel, A.~Carney \emph{et~al.}, ``Openai o1 system card,''
  \emph{arXiv preprint arXiv:2412.16720}, 2024.

\bibitem{o3-mini}
\BIBentryALTinterwordspacing
OpenAI, ``Openai o3-mini system card,'' January 2025, accessed: March 15, 2025.
  [Online]. Available: \url{https://openai.com/index/o3-mini-system-card/}
\BIBentrySTDinterwordspacing

\bibitem{sudhakaran2024mariogpt}
S.~Sudhakaran, M.~Gonz{\'a}lez-Duque, M.~Freiberger, C.~Glanois, E.~Najarro,
  and S.~Risi, ``Mariogpt: Open-ended text2level generation through large
  language models,'' \emph{Advances in Neural Information Processing Systems},
  vol.~36, 2024.

\bibitem{todd2023level}
G.~Todd, S.~Earle, M.~U. Nasir, M.~C. Green, and J.~Togelius, ``Level
  generation through large language models,'' in \emph{Foundations of Digital
  Games Conference}, 2023.

\end{thebibliography}
\bibliographystyle{IEEEtran}




\end{document}